\documentclass{bmvc2k}

\usepackage{graphicx}
\usepackage[normalem]{ulem}
\usepackage{epstopdf}
\usepackage{amsmath,amssymb} % define this before the line numbering.
\usepackage{color}

\usepackage[accsupp]{axessibility}  
\usepackage{diagbox}

\usepackage{subcaption}
\usepackage{tikz}
\usepackage{booktabs}
\usepackage{csquotes}
\usepackage{mathtools}
\usepackage{multirow}
\usepackage{enumitem} %< control
\usepackage{capt-of}% or \usepackage{caption}
\usepackage{color, colortbl}
\definecolor{LightCyan}{RGB}{241, 255, 231}
\definecolor{Col2}{RGB}{237, 242, 234}

\renewcommand{\paragraph}[1]{\vspace{1em}\noindent\textbf{#1}.}
\newcommand\Tau{\mathcal{T}}
\newcommand\Hau{\mathcal{H}}

\newcommand\Lau{\mathcal{L}}

\newcommand{\SC}[1]{\textcolor{red}{#1}}
\newtheorem{definition}{Definition}
\runninghead{Farkhondeh ET AL}{Towards Self-Supervised Gaze Estimation}

\begin{document}

%% Enter your paper number here for the review copy
%\bmvcreviewcopy{549}

\title{Towards Self-Supervised Gaze Estimation}

% Enter the paper's authors in order
% \addauthor{Name}{email/homepage}{INSTITUTION_CODE}
\addauthor{Arya Farkhondeh}{farkhondeh.1860768@studenti.uniroma1.it}{1,3}
\addauthor{Cristina Palmero}{crpalmec7@alumnes.ub.edu}{2,3}
\addauthor{Simone Scardapane}{simone.scardapane@uniroma1.it}{1}
\addauthor{Sergio Escalera}{sergio@maia.ub.es}{2,3}

% Enter the institutions
% \addinstitution{Name\\Address}
\addinstitution{
 Sapienza University of Rome\\
 Rome, Italy
}
\addinstitution{
 University of Barcelona\\
 Barcelona, Spain
}
\addinstitution{
 Computer Vision Center (CVC)\\
 Barcelona, Spain
}
\maketitle
\vspace{-0.3cm}
\begin{abstract}
Recent joint embedding-based self-supervised methods have surpassed standard supervised approaches on various image recognition tasks such as image classification. These self-supervised methods aim at maximizing agreement between features extracted from two differently transformed views of the same image, which results in learning an invariant representation with respect to appearance and geometric image transformations. However, the effectiveness of these approaches remains unclear in the context of gaze estimation, a structured regression task that requires equivariance under geometric transformations (e.g., rotations, horizontal flip). In this work, we propose SwAT, an equivariant version of the online clustering-based self-supervised approach SwAV, to learn more informative representations for gaze estimation. We demonstrate that SwAT, with ResNet-50 and supported with uncurated unlabeled face images, outperforms state-of-the-art gaze estimation methods and supervised baselines in various experiments. In particular, we achieve up to 57\% and 25\% improvements in cross-dataset and within-dataset evaluation tasks on existing benchmarks (ETH-XGaze, Gaze360, and MPIIFaceGaze). \vspace{-0.45cm}

\end{abstract}
\section{Introduction}
\label{sec:intro}
\vspace{-0.25cm}
Appearance-based gaze estimation remains a non-trivial problem to solve within the computer vision field due to the large variability across appearance and geometric factors. Convolutional neural network (CNN) based methods~\cite{7299081, Krafka2016EyeTF, Kellnhofer2019Gaze360PU, 9050633, Park2018DeepPG, Chen2018AppearanceBasedGE, Palmero2018RecurrentCF} have achieved promising performances fueled by large-scale datasets~\cite{Krafka2016EyeTF, Kellnhofer2019Gaze360PU, Zhang2020ETHXGaze}. Nonetheless, there is still a large gap to achieve a desirable performance especially when it comes to generalizing to unseen distributions with novel head poses, appearances, geometry, and illuminations. One way to address this problem is through the acquisition of even larger in-the-wild, gaze-annotated datasets with more variability. However, collecting data with accurate gaze annotations is an unscalable and laborious process that requires controlled conditions, complicated setups, tedious camera calibration, and subject recruitment. An inexpensive solution is therefore needed to extend variability in terms of appearance and geometric factors.

\begin{figure*}
  \centering
  \includegraphics[width=1.0\linewidth]{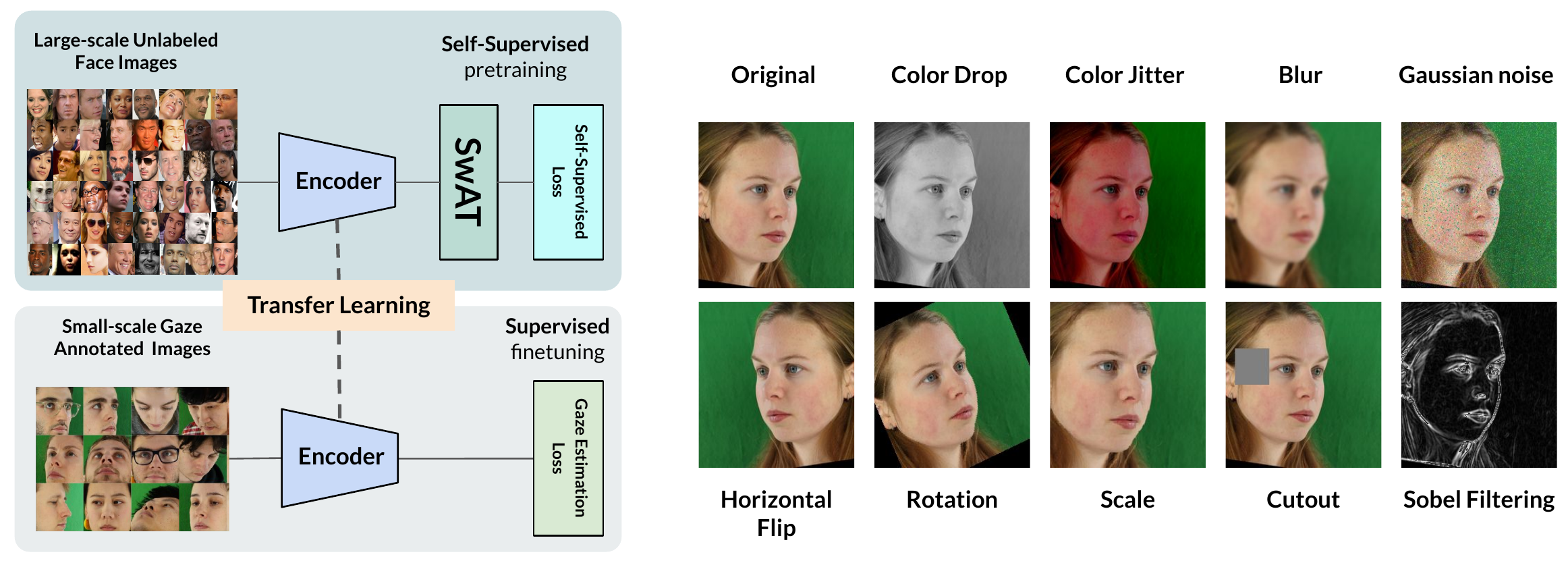} %\vspace{0.2cm}
   \caption{\textbf{Left. Global view of our approach. }In the first stage, we pretrain an encoder via an online clustering approach on a large-scale set of unlabeled face images while encouraging equivariance through our proposed method (SwAT). In the second stage, we transfer the learned knowledge from the first stage and fine-tune on a small-scale set of gaze-annotated images. \textbf{Right. Transformation Catalog. }The appearance and geometric transformations explored in this work for self-supervised representation learning.}
   \vspace{-0.5cm}
   \label{fig:app}
\end{figure*}

Recently, joint embedding-based self-supervised methods, including contrastive and non-contrastive, have obtained remarkable accuracy on various vision tasks, such as image classification~\cite{chen2020simple, Grill2020BootstrapYO, caron2020unsupervised, chen2020simsiam, Bardes2021VICRegVR}, object detection~\cite{xie2021detco}, and hand-pose estimation~\cite{spurr2021peclr}. These approaches have proven successful at learning generalizable features by leveraging large-scale unlabeled data~\cite{Jaiswal2020ASO, 9462394}. Similarly, these methods could leverage the vast amount of unlabeled face images that are publicly available on the Internet to learn useful representations for appearance-based gaze estimation. However, little attention has been paid to investigating their effectiveness for the gaze estimation task. Therefore, the main goal of this work is to explore the efficacy of a self-supervised approach in the context of gaze estimation to reduce the reliance on large-scale gaze-annotated data that is laborious to acquire. We specifically focus on full-face instead of eye-only images as input since the face provides auxiliary information~\cite{Krafka2016EyeTF, zhang2017s, Palmero2018RecurrentCF}.

In a nutshell, self-supervised learning aims at solving a pretext task to learn a useful representation. The representation is then used in downstream tasks via transfer learning. The common pretext task among (non-)contrastive self-supervised methods (e.g., SimCLR~\cite{chen2020simple}, MoCo~\cite{He2020MomentumCF}, SwAV~\cite{caron2020unsupervised}, BYOL~\cite{Grill2020BootstrapYO}, and VICReg~\cite{Bardes2021VICRegVR}) is to enforce consistency between features extracted from two differently transformed views of the same image. As a result, the feature extractor is encouraged to learn an invariant representation with respect to the image-space transformations, such as appearance (e.g., color jitter) and geometric (e.g., horizontal flip). Although invariance might be a desired property for most image recognition tasks, the structured regression task of gaze estimation requires equivariance under geometric transformations. In fact, applying geometric transformations to a face/eye image results in respective changes in gaze direction. Thus, in this work, our goal is to learn an equivariant representation under geometric transformations to align with the gaze estimation task. 

In this paper, we propose \textbf{Sw}apping \textbf{A}ffine \textbf{T}ransformations (\textbf{SwAT}), a novel method to achieve the desired property of equivariance. It can be thought of as a plug-and-play method that can be added to any joint embedding-based self-supervised approach. As Fig.~\ref{fig:app} depicts, we perform self-supervised pretraining on large-scale unlabeled face images while encouraging equivariance through SwAT. Then, we transfer the learned knowledge to the downstream gaze estimation task and finetune with gaze labels. Intuitively, SwAT allows the feature extractor to transfer the image-space geometric transformation to the representation output which preserves the intrinsic structure of the transformations. %\vspace{0.1cm}

Our proposed self-supervised approach potentially deconcentrates research in gaze estimation from the non-trivial process of large-scale annotated data collection towards effectively leveraging widely available large-scale unlabeled data. More importantly, leveraging such unlabeled data with more variety enhances the generalizability of gaze estimation models upon novel distributions. We show that the equivariance property provided by SwAT leads to learning better representations for gaze estimation, compared to other pretraining regimes. We also show that the unsupervised features provided by SwAT surpass the commonly used ImageNet supervised features in gaze estimation. We perform extensive experiments to verify the effectiveness of our approach under various challenging evaluation settings. We demonstrate that SwAT outperforms the supervised baselines in low-data regimes where only a few annotations (10\% and 30\%) are available. Supported with unlabeled data, SwAT achieves state-of-the-art results on existing benchmarks and improves the supervised baselines for cross- and within- dataset evaluation tasks by 57\% and 25\%, respectively. 
\vspace{-0.6cm}
\section{Related Work}
\label{sec:realtedwork}
\vspace{-0.25cm}
\textbf{Self-Supervised Learning. }Early self-supervised approaches attempted to learn useful representations from unlabeled data via solving handcrafted pretext tasks such as Jigsaw puzzle~\cite{Noroozi2016UnsupervisedLO},  colorization~\cite{zhang2016colorful}, transformation prediction~\cite{7410370, gidaris2018unsupervised}, and inpainting~\cite{Pathak2016ContextEF}. More recently, contrastive-based methods~\cite{He2020MomentumCF, chen2020simple, Misra2020SelfSupervisedLO} have achieved notable results on various computer vision tasks such as image classification. However, these methods are inherently computationally inefficient as they require pairwise contrasts with a large set of negative examples. Consequently, non-contrastive approaches~\cite{caron2020unsupervised, Grill2020BootstrapYO, chen2020simsiam, Bardes2021VICRegVR} are receiving special attention. Clustering-based approaches such as SwAV~\cite{caron2020unsupervised} discriminate between groups of images with similar features instead of individual images. However, both contrastive and non-contrastive methods are designed to learn invariant representations under image transformations, while gaze estimation requires equivariance under geometric transformations. Hence, in this work, we extend SwAV~\cite{caron2020unsupervised} via introducing equivariance under geometric transformations. Equivariance in self-supervised learning is starting to attract attention~\cite{spurr2021peclr, Dangovski2021EquivariantCL, Xie_2022_CVPR}. Despite their proven effectiveness, these methods bear some limitations that do not align with our assumptions. While our goal is to promote equivariance for multiple affine transformations, Dangovski et al.~\cite{Dangovski2021EquivariantCL}'s work is limited to a single transformation and Xie et al.~\cite{Xie_2022_CVPR}'s method is not scalable as the number of transformations increments. Most similarly, Spurr et al.~\cite{spurr2021peclr} propose an equivariance formulation for the task of 3D hand-pose estimation. However, their equivariance formulation together with a contrastive loss explicitly pushes apart the pseudo-negative pairs that may include faces with similar affine information, gaze, and head directions. %\vspace{0.1cm}

\noindent \textbf{Appearance-based Gaze Estimation. }Recent progress in appearance-based gaze estimation has been mainly achieved via collecting large-scale datasets~\cite{Mora2014EYEDIAPAD, Krafka2016EyeTF, Fischer2018RTGENERE, Kellnhofer2019Gaze360PU, Zhang2020ETHXGaze}, task-specific tailored architectures~\cite{Park2018DeepPG, Chen2018AppearanceBasedGE, Cheng2020ACA, D_2021_CVPR}, and data normalization methods~\cite{Zhang2018RevisitingDN,Zhang2019MPIIGazeRD}. Apart from supervised gaze estimation, weakly-supervised and unsupervised methods have started to receive more attention in gaze estimation. Kothari et al.~\cite{Kothari2021WeaklySupervisedPU} propose a weakly-supervised approach based on videos of people looking at each other. MTGLS~\cite{Ghosh2021MTGLSMG} utilizes off-the-shelf models to obtain pseudo labels for unlabeled eye images in order to learn a gaze representation. Recent generative-based unsupervised gaze estimation approaches~\cite{Yu2020UnsupervisedRL, Sun_2021_ICCV} make use of unlabeled eye images to learn gaze representations. Nevertheless, these approaches have limitations as they require supervision in the form of paired eye images of the same person~\cite{Yu2020UnsupervisedRL, Sun_2021_ICCV} with similar head-pose~\cite{Yu2020UnsupervisedRL}. Wu et al.~\cite{9717236} employ self-supervision as an auxiliary task for supervised gaze estimation. Unlike the previous methods, we pretrain a standard CNN architecture for gaze estimation in a self-supervised fashion via leveraging large-scale unlabeled face images. Our approach is less complex while more scalable as it does not make any assumption on the kind of unlabeled data and does not require multiple auxiliary losses for training as in~\cite{Yu2020UnsupervisedRL, Sun_2021_ICCV}. Furthermore, in contrast to previous unsupervised works that use eye images, we use full-face images, which have been proven to contain useful auxiliary information (e.g., head-pose, geometric features) for gaze estimation~\cite{Krafka2016EyeTF, zhang2017s, Palmero2018RecurrentCF}.
\vspace{-0.9cm}
\section{Method}
\label{sec:method}
\vspace{-0.25cm}
As Fig.~\ref{fig:app} depicts, our goal is to pretrain an encoder on large-scale unlabeled face images using a self-supervised approach (Sec.~\ref{subsec:ssl}) while encouraging equivariance via SwAT (Sec.~\ref{subsec:equ}).  Afterward, we transfer the knowledge to the gaze estimation task via supervised finetuning (Sec.~\ref{subsec:gazeest}). \vspace{-0.35cm}

\subsection{Self-Supervised Pretraining}
\label{subsec:ssl}
\vspace{-0.2cm}
\begin{figure*}[t]
  \centering
  \includegraphics[width=0.99\linewidth]{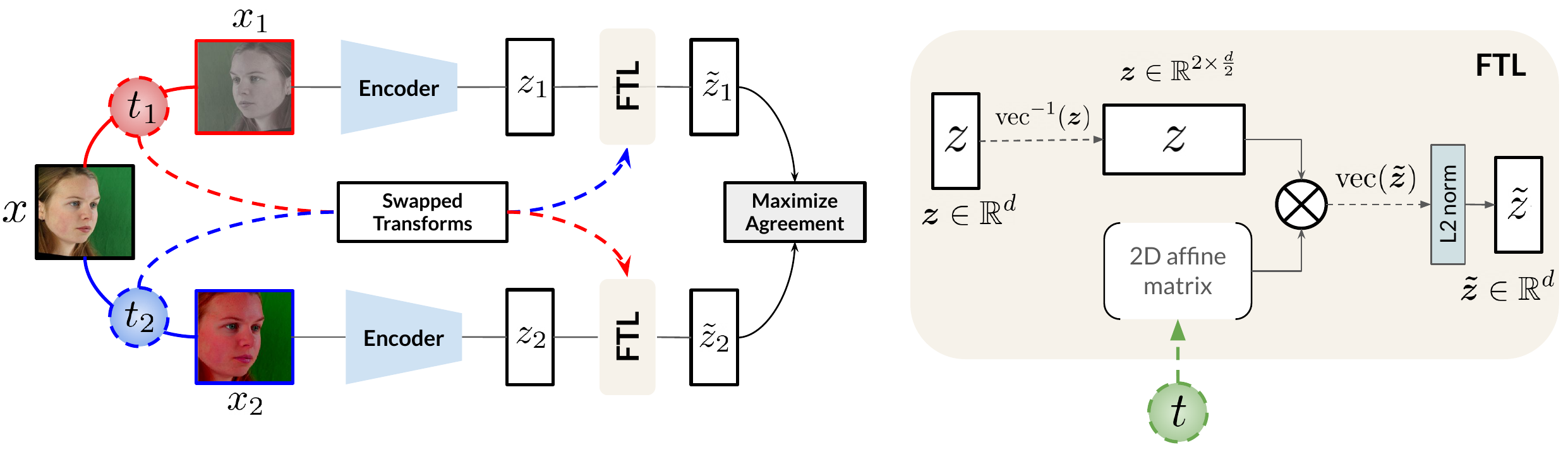} \vspace{-0.2cm}
   \caption{\textbf{Left. }Two differently transformed versions of the same image $\boldsymbol{x}$ are obtained via applying two different sets of transformations i.e., $\boldsymbol{x}_1 = t_{1}(\boldsymbol{x})$ and $\boldsymbol{x}_2 = t_{2}(\boldsymbol{x})$. An encoder is used to map the transformed views to vector representations, $\boldsymbol{z}_1$ and $\boldsymbol{z}_2$. To achieve equivariance, we equalize $\boldsymbol{z}_1$ and $\boldsymbol{z}_2$ in terms of affine information. To do so, we swap the affine transformations applied in image space $t_1$ and $t_2$, then we use the feature transform layer (FTL) to apply the swapped transformations to the feature vectors i.e., $\boldsymbol{\Tilde{z}}_1 = \mathbb{FTL} \big(t_{2}, \boldsymbol{z}_{1}\big)$ and $\boldsymbol{\Tilde{z}}_2 = \mathbb{FTL} \big(t_{1}, \boldsymbol{z}_{2}\big)$. Then, we maximize agreement between the resulting feature vectors, $\boldsymbol{\Tilde{z}}_1$ and $\boldsymbol{\Tilde{z}}_{2}$. \textbf{Right. }Details of the feature transform layer (FTL). $\textrm{vec}^{-1}(\boldsymbol{z})$ transforms $\boldsymbol{z}$ from 1D to 2D in order to enable matrix-matrix multiplication with the 2D affine matrix, resulting in $\boldsymbol{\Tilde{z}}$. Then, $\textrm{vec}(\boldsymbol{\Tilde{z}})$ transforms back $\boldsymbol{\Tilde{z}}$ from 2D to 1D. L2-norm is then applied. }
   \vspace{-0.4cm}
   \label{fig:method} 
\end{figure*}

%$R$ transforms $\boldsymbol{z}$ from 1D to 2D via reshaping $\boldsymbol{z} \in \mathbb{R}^{d}$ to $\boldsymbol{z} \in \mathbb{R}^{2 \times \frac{d}{2}}$, where $d$ is the dimensionality of the projection head. $\boldsymbol{\Tilde{z}}$ is transformed back to a 1D feature vector via $R^{-1}$.}

%Here, $\textrm{R}$ is a reshaping transformation (from $\boldsymbol{z} \in \mathbb{R}^{d}$ to $\boldsymbol{z} \in \mathbb{R}^{2 \times \frac{d}{2}}$) and $\textrm{R}^{-1}$ is its inverse transformation.}

%\AC{Here, $\textrm{vec}^{-1}(\boldsymbol{z})$ is the inverse of vectorization.}
In this work, as pretext task, we aim at maximizing the mutual information between the features from two different views of the same image. As shown in Fig.~\ref{fig:method} (left), two differently transformed views of an image $\boldsymbol{x}$ are computed via applying two different sets of transformations i.e., $\boldsymbol{x}_{1} = t_{1}(\boldsymbol{x})$ and $\boldsymbol{x}_2 = t_{2}(\boldsymbol{x})$ where, $t_{1} \sim \Tau$ and $t_{2} \sim \Tau$ are sampled from the same transformation catalog $\Tau$. An encoder $f_{\phi}(.)$ parameterized by $\phi$ maps the transformed views to vector representations, $\boldsymbol{z}_1 = f_{\phi}(\boldsymbol{x}_1)$ and $\boldsymbol{z}_2 = f_{\phi}(\boldsymbol{x}_2)$. 
The encoder $f_{\phi}(.)$ is composed of a backbone (e.g., ResNet) and a projection head (e.g., MLP). Then, we maximize agreement between the feature vectors using an online clustering-based self-supervised approach called SwAV~\cite{caron2020unsupervised}. SwAV enforces agreement using intermediate cluster assignments computed in an online fashion, where the cluster assignments are treated as the targets to predict from feature vectors. To compute the cluster assignments $\boldsymbol{c}_1$ and $\boldsymbol{c}_2$, the vector representations ($\boldsymbol{z}_1$ and $\boldsymbol{z}_2$) are compared to a set of M learnable prototype vectors $\boldsymbol{P}_{\psi} =\{\boldsymbol{p}_1, ..., \boldsymbol{p}_M\}$, parameterized by $\psi$. Maximizing agreement is achieved via swapping the computed cluster assignments and predicting them using feature vectors. The idea is to predict the cluster assignment $\boldsymbol{c}_1$ from the feature $\boldsymbol{z}_2$, and $\boldsymbol{c}_2$ from $\boldsymbol{z}_1$. Intuitively, if two feature vectors contain mutual information then it should be possible to predict the cluster assignment $\boldsymbol{c}_1$ ($\boldsymbol{c}_2$) from the other feature $\boldsymbol{z}_2$ ($\boldsymbol{z}_1$). The self-supervised loss function is as follows:
\vspace{-0.25cm}
\begin{equation}
\label{equ:swappred}
\Lau_{\textrm{SwAV}} = \ell (\boldsymbol{z}_1,\boldsymbol{c}_2) + \ell (\boldsymbol{z}_2,\boldsymbol{c}_1),
\end{equation}  
\vspace{-0.6cm}

\noindent where $\ell (\boldsymbol{z},\boldsymbol{c})$ is the cross entropy loss between the cluster assignments and the probability computed by applying softmax to the dot products of $\boldsymbol{z}_i$ and prototypes ($\boldsymbol{P}_{\psi}$), as in Eq.~\ref{equ:entropy}. The cross entropy loss measures agreement between a feature and cluster assignment. $\ell (\boldsymbol{z}_i,\boldsymbol{c}_j)$ is defined as follows:
\vspace{-0.2cm}
\begin{equation}
\label{equ:entropy}
\ell (\boldsymbol{z}_i,\boldsymbol{c}_j) = - \sum_{m} \boldsymbol{c}_{j}^{(m)} \log \Bigg(\frac{\exp (\frac{1}{\tau} \boldsymbol{z}{_i}^{\top} \boldsymbol{p}_{m})}{\sum_{m^{'}} \exp (\frac{1}{\tau} \boldsymbol{z}_{i}^{\top} \boldsymbol{p}_{m^{'}})}\Bigg),
\end{equation} %\vspace{0.2cm}

\noindent where $\tau$ is a temperature parameter and $m$ denotes the $m$th prototype. The overall loss function (Eq.~\ref{equ:swappred}) is minimized with respect to both parameters of the encoder $\phi$ and trainable prototypes $\psi$. The method is online since only the features within a batch are used to compute the cluster assignments. To avoid trivial solutions i.e., assigning the same cluster for every image within a batch, score adjustment is performed using an optimal transport algorithm, namely  Sinkhorn-Knopp~\cite{Sinkhorn}. It encourages equipartition guaranteeing that the cluster assignments are distinct for images within a batch. \vspace{-0.3cm}

\subsection{Equivariant Representation Learning}
\label{subsec:equ}
\vspace{-0.2cm}
Similar to other (non-)contrastive self-supervised approaches, the SwAV formulation (Sec.~\ref{subsec:ssl}) encourages invariance under appearance and geometric transformations. In image recognition tasks such as image classification, applying geometric transformations ($t^g$) to an image does not change the label. However, in the gaze estimation task, applying geometric transformations in image space results in respective changes in label space. Thus, instead of learning an \textit{invariant} representation, we aim at learning an \textit{equivariant} representation.

\vspace{-0.15cm}
\begin{definition}[Equivariance]
\label{def:equ}
A mapping function $f_{\phi}: \boldsymbol{x} \rightarrow \boldsymbol{z}$ is said to be equivariant with respect to image-space transformation $t^{g}_{I}$ when mapping the transformed input image, $f_{\phi}(t^{g}_{I} (\boldsymbol{x}))$, produces the same result as transforming the vector representation of the input image, i.e., $t^{g}_{F}(f_{\phi}(\boldsymbol{x}))$: 
\end{definition}
\vspace{-0.3cm}
\begin{equation}
\label{equ:eq}
f_{\phi}(t^{g}_{I} (\boldsymbol{x})) = t^{g}_{F}(f_{\phi}(\boldsymbol{x})),
\end{equation}\vspace{-0.4cm}

\noindent where transformations $t^{g}_{I}$ and $t^{g}_{F}$ are used to apply the same transformation in different spaces i.e., image space and feature space, respectively. Intuitively, the \textit{equivariance} property enables $f_{\phi}$ to learn a direct relationship between image space and feature space, thereby preserving the intrinsic structure of the transformations~\cite{Worrall2017InterpretableTW}. \vspace{0.1cm}

\noindent \textbf{Swapping Affine Transformations. }Eq.~\ref{equ:swappred} enforces consistent mapping between two transformed views via intermediate cluster assignments. Abstractly, it aims to maximize the mutual information between the features from two views. Thus, ideally, \vspace{-0.3cm}

\begin{equation}
\label{equ:zz}
f_{\phi} (t^{g}_{1} (\boldsymbol{x})) = f_{\phi} (t^{g}_{2} (\boldsymbol{x})).
\end{equation} \vspace{-0.4cm}

The only way that the above equality is satisfied is through encouraging $f_{\phi}$ to be invariant with respect to the applied geometric transformations $t^{g}_{1}$ and $t^{g}_{2}$. Instead, to let the mapping function $f_{\phi}$ be equivariant under affine transformations applied in image space, we propose the  \textbf{Sw}apping \textbf{A}ffine \textbf{T}ransformations (\textbf{SwAT}) method. SwAT achieves equivariance via equalization of vector representations in terms of applied image-space affine transformations. To achieve that, as in Eq.~\ref{equ:applyftl} and Fig.~\ref{fig:method}, we swap the affine transformations applied in image-space, and then we apply them in feature-space via a feature transform layer ($\mathbb{FTL}$), detailed later. Thus, \vspace{-0.5cm}

\begin{equation}
\label{equ:applyftl}
\boldsymbol{\Tilde{z}}_{1} = \mathbb{FTL} \big(t^{g}_{2}, \boldsymbol{z}_{1}\big), \hspace{0.3cm} \boldsymbol{\Tilde{z}}_{2} = \mathbb{FTL} \big(t^{g}_{1}, \boldsymbol{z}_{2}\big).
\end{equation} \vspace{-0.5cm}

Intuitively, $\boldsymbol{\Tilde{z}}_{1}$ and $\boldsymbol{\Tilde{z}}_{2}$ contain the same affine transformation information. Thus, enforcing consistency between transformation-equalized vector representations prevents $f_{\phi}$ from becoming invariant with respect to transformations. In contrast, since unequalized vector representations $\boldsymbol{z}_{1}$ and $\boldsymbol{z}_{2}$ contain different transformation information, enforcing consistency would result in invariance as in Eq.~\ref{equ:zz}. The self-supervised loss (Eq.~\ref{equ:swappred}) becomes:
\vspace{-0.1cm}
\begin{equation}
\label{equ:swapequ}
\Lau_{\textrm{SwAT}} = \ell \big(\boldsymbol{\Tilde{z}}_{1}, \boldsymbol{\Tilde{c}}_{2}\big) + \ell \big(\boldsymbol{\Tilde{z}}_{2}, \boldsymbol{\Tilde{c}}_{1}\big) ,
\vspace{-0.2cm}
\end{equation}
\noindent where, %\vspace{-0.1cm}
\begin{equation}
\boldsymbol{\Tilde{c}}_{2} = \boldsymbol{\Tilde{z}}_{2} \hspace{0.1cm} \boldsymbol{P}_{\psi}, \hspace{0.5cm} \boldsymbol{\Tilde{z}}_{2} \in \mathbb{R}^{d}, \hspace{0.5cm} \boldsymbol{P}_{\psi} \in \mathbb{R}^{d \times \mathrm{M}}.
\end{equation} 

\noindent \textbf{Feature Transform Layer. }As Fig.~\ref{fig:method} (right) depicts, to be able to apply the feature-space equivalent ($t^{g}_{F}$) of the image-space transformation ($t^{g}_{I}$), we introduce a non-trainable feature transform layer ($\mathbb{FTL}$). This layer takes as input the 2D affine transformation matrix $\boldsymbol{T}_{\theta}$ (e.g., 2D rotation matrix with angle $\theta$) and 1D feature vector $\boldsymbol{z}$. It first transforms $\boldsymbol{z}$ from 1D to 2D via an inverse vectorization, $\textrm{vec}_{2 \times k}^{-1}(\boldsymbol{z})$, where $k=\frac{d}{2}$ and $d$ is the dimensionality of the projection head. Afterward, it performs a matrix-matrix multiplication, resulting in $\boldsymbol{\Tilde{z}}$. Finally, $\boldsymbol{\Tilde{z}}$ is transformed back to a 1D feature vector via a vectorization, $\textrm{vec}(\boldsymbol{\Tilde{z}})$, and then L2-norm is applied. \vspace{0.1cm}

\noindent \textbf{Transformations. }Fig.~\ref{fig:app} (right) shows the explored transformations in this work which fall into two groups, namely appearance and geometric transformations. In the context of gaze estimation, appearance and scale transformations do not change the 3D gaze direction label with respect to the camera coordinate system. In contrast, applying horizontal flip and rotation in image space results in respective changes in label space. Thus, for our proposed SwAT method, we only swap horizontal flip and rotation transformations. Further details of transformations can be found in the supplemental material (Sec.~\SC{B.2}). \vspace{-0.3cm}

\subsection{Finetuning for Gaze Estimation}
\label{subsec:gazeest}
\vspace{-0.2cm}
Gaze estimation is a regression task where the goal is to learn a mapping function $\Hau:\boldsymbol{x} \rightarrow \boldsymbol{g} $ that maps the high-dimensional RGB images $\boldsymbol{x}\in \mathbb{R}^{H\times W \times 3}$ to low-dimensional 2D angles $\boldsymbol{g} \in \mathbb{R}^{2}$ i.e., yaw and pitch. The 2D angles are a compact representation of the 3D gaze direction vector in the camera coordinate system, the origin of which is the center of the face or the midpoint between the eyes, depending on the dataset. $\Hau$ is a parameterized function, composed of a backbone encoder (e.g., ResNet) as well as a linear head (e.g., MLP). To perform gaze estimation, we first initialize the weights of the backbone with the pretrained weights previously learned through self-supervised pretraining. Then, the whole network is finetuned with gaze-annotated data using the L1 loss between the estimated angles $\boldsymbol{\hat{g}} = \Hau (x)$ and actual angles $g$, as follows (where $N$ is the number of samples): \vspace{-0.3cm}
\begin{equation}
\Lau_{\textrm{gaze}} = \frac{1}{N} \sum_{i=1}^{N} || \boldsymbol{g}_{i} - \boldsymbol{\hat{g}}_{i} ||_{1},
\label{equ:gaze}
\end{equation} \vspace{-0.9cm}

\section{Experiments and Results}
\label{sec:expres}
%\vspace{-0.25cm}
In this section, we assess the performance of the proposed SwAT method through an exhaustive experimental evaluation to demonstrate the utility of the equivariance property under different scenarios. We refer the reader to supplemental material for robustness analysis (Sec.~\SC{C}), ablation studies and comparison with~\cite{spurr2021peclr} (Sec.~\SC{D}), and qualitative results (Sec.~\SC{E}). %\vspace{-0.75cm}

\subsection{Experimental Setting}
\vspace{-0.15cm}
\noindent \textbf{Datasets. }For the self-supervised pretraining stage, we use a curated dataset i.e., ETH-XGaze~\cite{Zhang2020ETHXGaze} without labels. It contains 756,540 images and 80 subjects for training, captured under controlled laboratory conditions. Since ETH-XGaze was specifically collected for the task of gaze estimation under controlled conditions, it is unclear whether the quality of unsupervised features remains the same while using an uncurated dataset. To shed light on this, we also use the VGG-Face dataset~\cite{Parkhi15} for pretraining. VGG-Face is collected from the web, including 2,622 identities and about 1.5 M face images. For the finetuning phase, throughout various experiments, we use the publicly available Gaze360~\cite{Kellnhofer2019Gaze360PU} and MPIIFaceGaze~\cite{zhang2017s} datasets, in addition to ETH-XGaze. Gaze360 is a physically unconstrained dataset collected in indoor and outdoor environments with a wide range of head poses. MPIIFaceGaze is a subset of the MPIIGaze~\cite{7299081} dataset, recorded while doing activities on the laptop. \vspace{0.1cm}

\noindent \textbf{Implementation Details. }For the pretraining phase, we use SGD + LARS~\cite{You2017LargeBT} optimizer with a batch size of 1024 distributed over 8 NVIDIA GeForce RTX 3090 GPUs. We pretrain for 100 epochs and experimentally found it to be sufficient. We use a weight decay of $10^{-6}$ and the learning rate is set to 0.45 followed by an initial linear warmup stage for 10 epochs. Afterward, we use cosine learning rate decay~\cite{Loshchilov2017SGDRSG} with a final value of 0.00045. As the encoder, we use ResNet~\cite{He2016DeepRL} and a projection head that consists of a 2-layer MLP that maps the encoder output to 256-D. We experimentally set the number of prototypes $M$ to 500. We perform the finetuning stage for 100 epochs using Adam optimizer~\cite{adam-kingma}, with a batch size of 512. We decay the learning rate at 40 and 80 epochs by 0.1. We set the input size to 224 $\times$ 224 unless otherwise stated. Full details can be found in the supplemental material (Sec.~\SC{A}). \vspace{-0.3cm}

\noindent \textbf{Experimental protocol. }We use the dataset partitions provided by each dataset. A prior data normalization stage is commonly applied by creating a virtual camera with fixed intrinsic and extrinsic camera parameters, which reduces head pose variability and hence the training space~\cite{Zhang2018RevisitingDN}. However, this normalization may conceal the benefits of enforcing equivariance for geometric transformations, especially for already constrained datasets with little geometric variability. Furthermore, this stage cannot be applied accurately if camera parameters are not provided. Therefore, for the finetuning part of our methods (baselines and SwAT) we apply data normalization only to ETH-XGaze, since its test evaluation assumes normalized data, and to MPIIFaceGaze, to compare against previous approaches that performed the normalization stage. We also evaluate the unnormalized version of MPIIFaceGaze (referred to as MPIIFaceGaze\mbox{*}) to better quantify the benefits of SwAT and compare its performance against the normalized counterpart. Throughout the paper, we use average angular gaze error in degrees to measure performance. \vspace{-0.45cm} 

\subsection{Evaluating the Unsupervised Features}
\label{sec:unsupfeatt}
\vspace{-0.2cm}

After assessing the effectiveness of each individual transformation and finding an optimal composition for SwAV and SwAT (Sec.~\SC{B.1}, supplemental material), we evaluate the quality of unsupervised features. More precisely, the goal of this experiment is twofold: to explore whether the equivariance property provided by SwA\textbf{T} leads to a better representation compared to the invariance counterpart (SwA\textbf{V}), and to shed light on the quality of the unsupervised features with the curated (ETH-XGaze) and uncurated datasets (VGG-Face), used for pretraining. To do so, we perform a linear evaluation, where we freeze the backbone (ResNet-50) after pretraining and train a linear gaze regressor on top. Then, we measure the performance on the validation set that we manually create by splitting the available ETH-XGaze training set intro training and validation sets. We also compare the unsupervised features with ImageNet supervised features, which are widely used in current gaze estimation works as initialization. 

\begin{figure*}[t]
\begin{subfigure}{.5\textwidth}
  \centering
  \includegraphics[width=.99\linewidth]{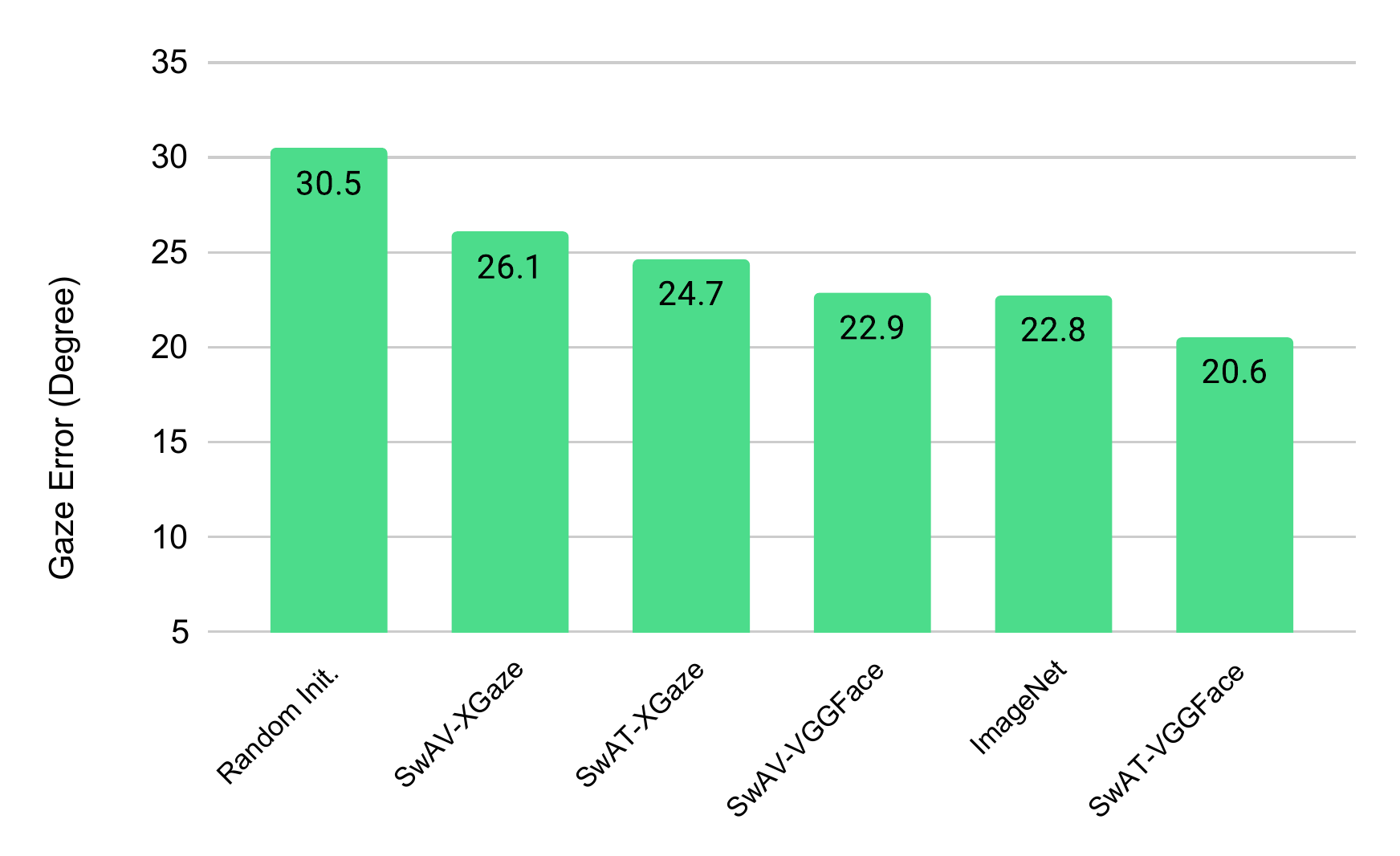}
\end{subfigure}%
\begin{subfigure}{.5\textwidth}
  \centering
  \includegraphics[width=.99\linewidth]{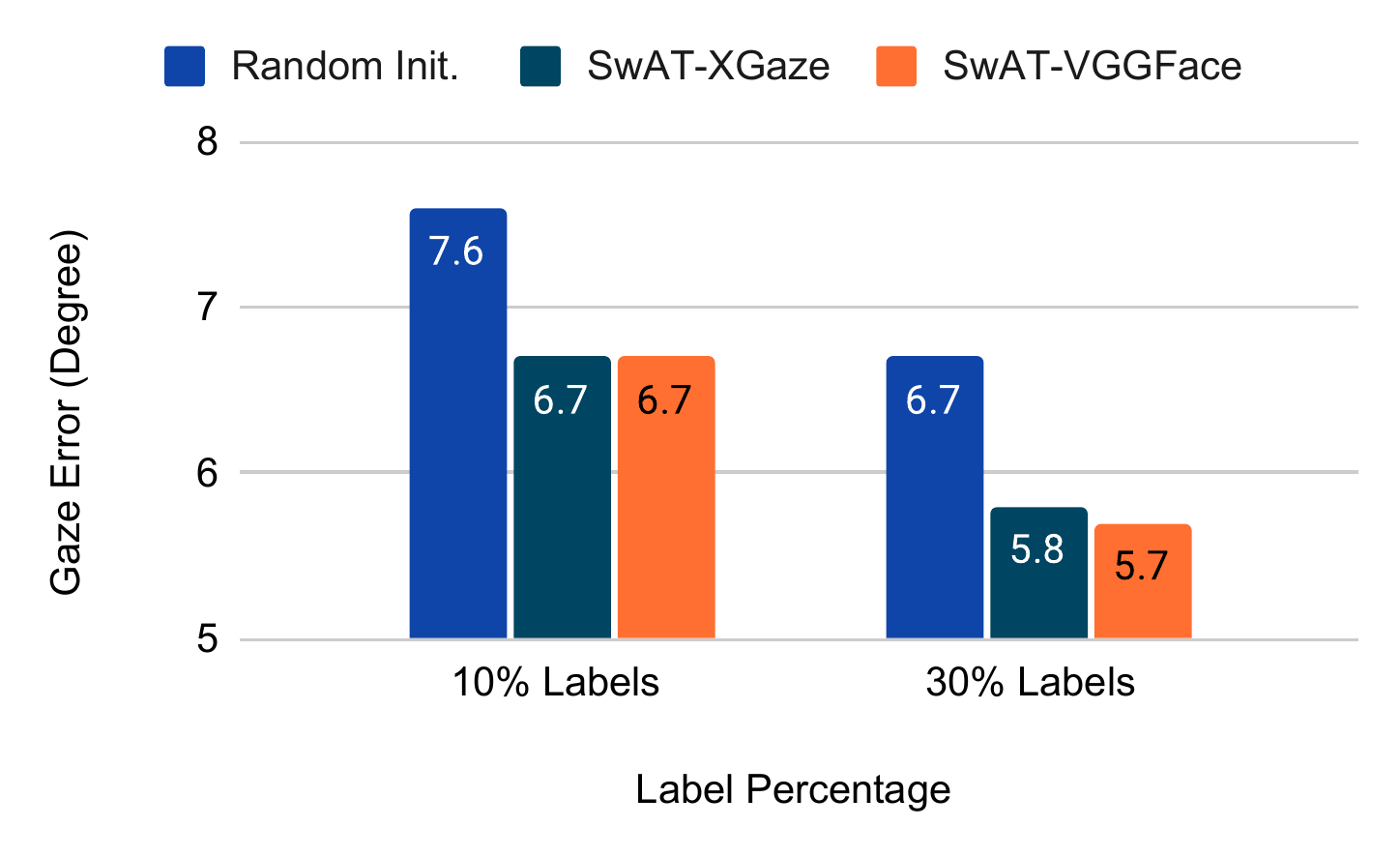}
\end{subfigure}
\caption{\textbf{Left}. Results of evaluating the unsupervised features of SwAV and SwAT pretrained with ETH-XGaze and VGG-Face datasets compared to random and ImageNet-based initializations. Performance is measured on the manually created validation set of ETH-XGaze.  \textbf{Right. }Results of semi-supervised learning using two subsets (10\% and 30\%) of the ETH-XGaze dataset at the subject level on the test set of ETH-XGaze.}
\vspace{-0.5cm}
\label{fig:unsemi}
\end{figure*}

Fig.~\ref{fig:unsemi} (left) shows the results of the linear evaluation on the validation set of ETH-XGaze. We can see that SwAT outperforms SwAV with both curated (ETH-XGaze) and uncurated (VGG-Face) datasets. More importantly, SwAT surpasses the supervised features pretrained on ImageNet, decreasing the gaze error from $22.8^{\circ}$ to $20.6^{\circ}$. In the next experiments, we focus on comparing and evaluating SwAT in presence of labels for finetuning. \vspace{-0.4cm}
\subsection{Semi-supervised Learning}
\label{sec:semisupp}
%\vspace{-0.2cm}
In this evaluation, we examine the label-efficiency of SwAT. To achieve that, we perform semi-supervised learning on two subsets of the ETH-XGaze dataset. More precisely, we define two subsets i.e., 10\% and 30\% at subject level, and finetune the whole network on these subsets. As a baseline, we train a counterpart on the same subsets and with the same architecture but instead of using pretrained SwAT weights, we randomly initialize the weights. Fig.~\ref{fig:unsemi} (right) depicts the results of the semi-supervised learning. As can be seen, ResNet-50 pretrained with SwAT improves the baseline up to $1.0^{\circ}$ when only 10\% and 30\% of labeled data at the subject level is available. This is of great importance in the gaze estimation context as recruiting fewer subjects saves cost and time. \vspace{-0.25cm}

\subsection{Comparison to state of the art}
\label{sec:sota}
%\vspace{-0.2cm}
\begin{table*}[t]
\centering
\resizebox{\linewidth}{!}{
\begin{tabular}{@{}lcccccc@{}}
\toprule
\textbf{Method} & \textbf{Pretrain} & \textbf{Arch.}  &\textbf{ETH-XGaze} & \textbf{Gaze360} & \textbf{MPIIFace} & \textbf{MPIIFace$^{*}$} \\
\midrule
Full-Face~\cite{zhang2017s} & ImageNet & AlexNet+SW & N/A & N/A & \textbf{4.8}  & N/A\\
Dilated-Net~\cite{Chen2018AppearanceBasedGE} & ImageNet & Dilated-CNN & N/A & N/A &  \textbf{4.8} & N/A\\
RT-GENE~\cite{Fischer2018RTGENERE} & ImageNet & VGG-16 & N/A & N/A &  \textbf{4.8} & N/A\\
Gaze360 \cite{Kellnhofer2019Gaze360PU} & ImageNet & ResNet-18&  N/A & 13.2 &  N/A & N/A\\
MTGLS \cite{Ghosh2021MTGLSMG} & MS-Celeb-1M & ResNet-50 & N/A & 12.8 & N/A & N/A\\
ETH-XGaze~\cite{Zhang2020ETHXGaze} & ImageNet & ResNet-50 & 4.5 & N/A & \textbf{4.8} & 7.1$^{\dagger}$\\
Wu et al.~\cite{9717236} & N/S & ResNet-18 & N/A & 13.2 &  N/A &  N/A\\
\midrule
%\rowcolor{Col2}
Baseline (ours) & Random Init. & ResNet-50& 5.9 & 12.2 &  5.7 & 8.5\\
\midrule
\rowcolor{LightCyan}
SwAT (ours) & ETH-XGaze  & ResNet-50& 4.5 & 11.9 &  5.2 & 7.5 \\
\rowcolor{LightCyan}
SwAT (ours) & VGG-Face & ResNet-50& \textbf{4.4} & \textbf{11.6} & 5.0 &  \textbf{6.9}\\
\bottomrule
\end{tabular}} \vspace{0.000005cm}
\caption{
Comparison of SwAT with state-of-the-art full-face appearance-based gaze estimation works, reported as average angular gaze error (degrees). Best results are bolded. Performances of the state-of-the-art approaches are shown as reported by their authors, except values marked with ${}^{\dagger}$. MPIIFaceGaze$^{*}$ denotes the unnormalized version of MPIIFaceGaze.
} \vspace{-0.1cm}
\label{table:sota}
\end{table*}
\begin{table*}[t]
\centering%\scriptsize
%\resizebox{\linewidth}{!}{
%\renewcommand\arraystretch{0.5}
\footnotesize
%\resizebox{\linewidth}{!}{
\begin{tabular}{@{}llccccc@{}}
\toprule
\textbf{Method} &\backslashbox{\textbf{Train}}{\textbf{Test}} & ETH-XGaze & Gaze360 & MPIIFace & MPIIFace$^{*}$\\
\midrule
\multirow{4}{6em}{} & ETH-XGaze & - & 30.0 & 23.5 & 17.5\\
Supervised &Gaze360 & 25.6 &-&  30.4 & 21.5\\
&MPIIFace & 32.2 & 27.4 & -  &  -\\
&MPIIFace$^{*}$ & 35.5 & 28.9 &  - & -\\
\midrule
\rowcolor{LightCyan}
\multirow{4}{6em}{SwAT} & ETH-XGaze& - & \textbf{22.9} & \textbf{12.1} & \textbf{11.6}\\
\rowcolor{LightCyan}
SwAT &Gaze360 & \textbf{19.4} &-& \textbf{13.0} & \textbf{12.8}\\
\rowcolor{LightCyan}
&MPIIFace & \textbf{29.5} & \textbf{24.9} & - & -\\
\rowcolor{LightCyan}
&MPIIFace$^{*}$ & \textbf{32.6}  & \textbf{25.5} & -  &   -\\
\bottomrule
\end{tabular}
\vspace{0.2cm}
\caption{
Comparison between supervised baseline and SwAT on cross-dataset evaluation. Numbers denote gaze error in degrees. Best results are bolded.
} 
\vspace{-0.3cm}
\label{table:crosseval}
\end{table*}
We compare SwAT with state-of-the-art methods for full-face appearance-based gaze estimation. We pretrain SwAT with ResNet-50 as encoder on ETH-XGaze (without labels) and VGG-Face datasets. Then, we finetune the whole network using the aforementioned datasets. As a baseline, we also train the same encoder (ResNet-50) solely in a supervised fashion. Tab.~\ref{table:sota} shows the comparison with the state of the art along with the datasets used for pretraining and the type of encoder. As can be seen, the supervised baseline is unable to outperform the state of the art, except on Gaze360. However, the same encoder boosted with SwAT unsupervised pretrained features achieves up to 25\%, 5\%, 14\%, and  19\% improvements compared to the supervised baseline on ETH-XGaze, Gaze360, MPIIFaceGaze, and MPIIFaceGaze$^{*}$, respectively. Furthermore, SwAT pretrained with the VGG-Face dataset outperforms SwAT pretrained on ETH-XGaze (without labels) on all four benchmarks. This suggests that SwAT can effectively make use of uncurated datasets. On ETH-XGaze, SwAT pretrained with VGG-Face outperforms the state of the art that utilizes the pretrained ImageNet supervised weights. In addition, SwAT improves the state of the art up to 9\% on Gaze360 while slightly underperforming it on MPIIFaceGaze. However, we can better observe the benefit of SwAT on the unnormalized version of MPIIFaceGaze (MPIIFaceGaze$^{*}$), where SwAT improves the ETH-XGaze method with no data normalization by 0.2$^{\circ}$. These results demonstrate the superior performance of SwAT in unrestricted scenarios. \vspace{-0.3cm} 

\subsection{Cross-dataset Evaluation}
\label{sec:crossdata}
\vspace{-0.2cm}
To evaluate the out-of-distribution generalization capability of SwAT, we perform a cross-dataset evaluation, i.e., training on a given dataset and testing on other datasets. We consider four datasets, namely, ETH-XGaze, Gaze360, MPIIFaceGaze, and MPIIFaceGaze$^{*}$. We use ResNet-50 as encoder, pretrained on VGG-Face using SwAT. We compare our self-supervised approach (SwAT) to a supervised baseline that is solely trained in a supervised fashion. Tab.~\ref{table:crosseval} shows the results of cross-dataset evaluation. SwAT improves the supervised baseline by a large amount.  In detail, SwAT achieves up to 24\% relative improvement on the ETH-XGaze dataset, and outperforms the supervised counterpart by 24\% on Gaze360, by 57\% on MPIIFaceGaze, and by 41\% on MPIIFaceGaze$^{*}$. 

%\vspace{-0.4cm}
\subsection{Equivariance Analysis}
\label{sec:equana}
%\vspace{-0.2cm}
\begin{figure*}[t]
\begin{subfigure}{.5\textwidth}
  \centering
  \includegraphics[width=.99\linewidth]{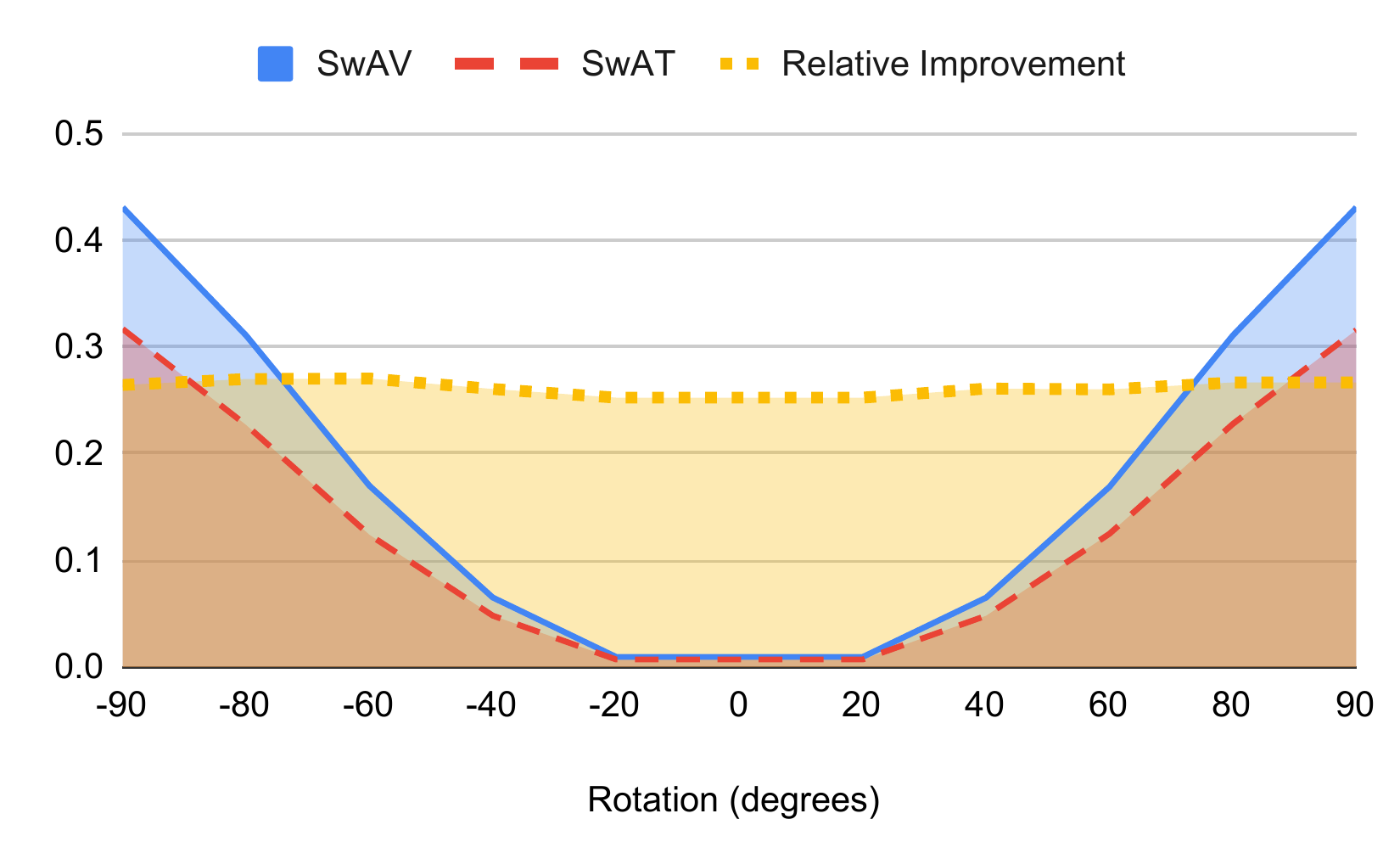}
\end{subfigure}%
\begin{subfigure}{.5\textwidth}
  \centering
  \includegraphics[width=.99\linewidth]{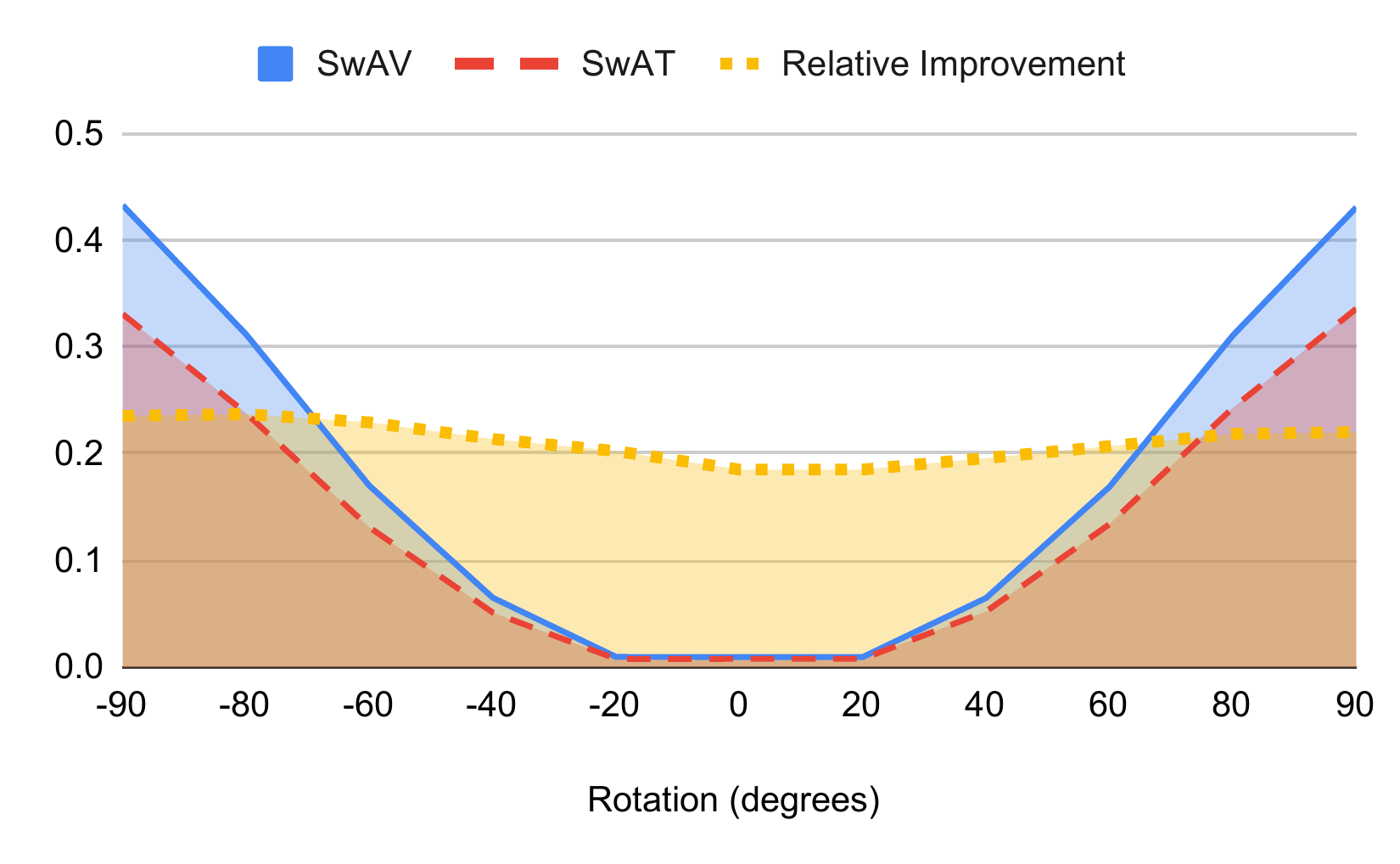}
\end{subfigure}\vspace{0.1cm}
  \vspace{-0.4cm}
   \caption{Results of calculating $\Lau_{equ}$ for SwAV and SwAT on Gaze360 (\textbf{Left}) and MPIIFaceGaze$^{*}$ (\textbf{Right}) datasets. The dotted lines shows the relative improvement achieved by SwAT over SwAV.}
   %\vspace{-0.55cm}
   \label{fig:eq}
\end{figure*}

To evaluate the equivariance capability, we rely on the definition of equivariance (Eq.~\ref{equ:eq}) and calculate the following metric ($\Lau_{equ}$): 

\begin{equation}
\label{equ:equv}
\Lau_{equ} = \frac{1}{N} \sum_{i=1}^{N} || f_{\phi}(t^{g}_{I} (\boldsymbol{x_{i}})) - t^{g}_{F}(f_{\phi}(\boldsymbol{x_{i}})) ||_{2}.
\end{equation}

\noindent We compare $f_{\phi}$ pretrained with SwAV and SwAT on the VGG-Face dataset. As the evaluation datasets, we specifically focus on unconstrained gaze scenarios and calculate $\Lau_{equ}$ for Gaze360 and MPIIFaceGaze$^{*}$. We expect SwAT to achieve lower values, which indicates enforcing equivariance. Fig.~\ref{fig:eq} depicts the results of $\Lau_{equ}$ on Gaze360 (left) and MPIIFaceGaze$^{*}$ (right), varying rotation degrees.  As shown, in both cases SwAT consistently outperforms SwAV in the whole rotation range. More precisely, on average, SwAT achieves 27\% and 21\% relative improvements compared to SwAV on Gaze360 and MPIIFaceGaze$^{*}$, respectively. Moreover, we calculate $\Lau_{equ}$ for horizontal flip and find that SwAT improves SwAV by 26\% on Gaze360 and 21\% on MPIIFaceGaze$^{*}$. 
\vspace{-0.4cm}
\section{Conclusion}
\vspace{-0.2cm}
In this paper, we explored the effectiveness of a self-supervised method in the context of gaze estimation, and proposed a novel approach (SwAT) to learn an equivariant representation for geometric transformations, i.e., rotations and horizontal flip. Our approach is task-agnostic and can be applied to any joint embedding-based self-supervised approach. We showed that SwAT learns more informative representations than other pretraining schemes for the task of gaze estimation. Our approach fueled by a large-scale uncurated dataset achieves more generalizable results, outperforming the supervised baselines and state-of-the-art approaches for both within- and cross-dataset settings. We also showed that our method achieves superior performance with fewer subjects. Thus, our approach can be leveraged to boost the performance of current gaze estimation systems in the real world via leveraging large-scale freely available face images on the Internet. 

\vspace{0.2cm}

\noindent \textbf{Acknowledgements}
\vspace{0.1cm}

\noindent This work has been partially supported by the Spanish project PID2019-105093GB-I00 and by ICREA under the ICREA Academia programme.
\bibliography{bibmain}

\begin{thebibliography}{45}
\providecommand{\natexlab}[1]{#1}
\providecommand{\url}[1]{\texttt{#1}}
\expandafter\ifx\csname urlstyle\endcsname\relax
  \providecommand{\doi}[1]{doi: #1}\else
  \providecommand{\doi}{doi: \begingroup \urlstyle{rm}\Url}\fi

\bibitem[Agrawal et~al.(2015)Agrawal, Carreira, and Malik]{7410370}
Pulkit Agrawal, João Carreira, and Jitendra Malik.
\newblock Learning to see by moving.
\newblock In \emph{ICCV}, 2015.

\bibitem[Bardes et~al.(2022)Bardes, Ponce, and LeCun]{Bardes2021VICRegVR}
Adrien Bardes, Jean Ponce, and Yann LeCun.
\newblock {VICR}eg: Variance-invariance-covariance regularization for
  self-supervised learning.
\newblock In \emph{ICLR}, 2022.

\bibitem[Caron et~al.(2020)Caron, Misra, Mairal, Goyal, Bojanowski, and
  Joulin]{caron2020unsupervised}
Mathilde Caron, Ishan Misra, Julien Mairal, Priya Goyal, Piotr Bojanowski, and
  Armand Joulin.
\newblock Unsupervised learning of visual features by contrasting cluster
  assignments.
\newblock In \emph{NeurIPS}, 2020.

\bibitem[Chen et~al.(2020)Chen, Kornblith, Norouzi, and Hinton]{chen2020simple}
Ting Chen, Simon Kornblith, Mohammad Norouzi, and Geoffrey Hinton.
\newblock A simple framework for contrastive learning of visual
  representations.
\newblock In \emph{ICML}, 2020.

\bibitem[Chen and He(2021)]{chen2020simsiam}
Xinlei Chen and Kaiming He.
\newblock Exploring simple siamese representation learning.
\newblock In \emph{CVPR}, 2021.

\bibitem[Chen and Shi(2018)]{Chen2018AppearanceBasedGE}
Zhaokang Chen and Bertram~E. Shi.
\newblock Appearance-based gaze estimation using dilated-convolutions.
\newblock In \emph{ACCV}, 2018.

\bibitem[Cheng et~al.(2020{\natexlab{a}})Cheng, Huang, Wang, Qian, and
  Lu]{Cheng2020ACA}
Yihua Cheng, Shiyao Huang, Fei Wang, Chen Qian, and Feng Lu.
\newblock A coarse-to-fine adaptive network for appearance-based gaze
  estimation.
\newblock In \emph{AAAI}, 2020{\natexlab{a}}.

\bibitem[Cheng et~al.(2020{\natexlab{b}})Cheng, Zhang, Lu, and Sato]{9050633}
Yihua Cheng, Xucong Zhang, Feng Lu, and Yoichi Sato.
\newblock Gaze estimation by exploring two-eye asymmetry.
\newblock In \emph{IEEE Transactions on Image Processing}, 2020{\natexlab{b}}.

\bibitem[Cuturi(2013)]{Sinkhorn}
Marco Cuturi.
\newblock Sinkhorn distances: Lightspeed computation of optimal transport.
\newblock In \emph{NeurIPS}, 2013.

\bibitem[D and Biswas(2021)]{D_2021_CVPR}
Murthy L~R D and Pradipta Biswas.
\newblock Appearance-based gaze estimation using attention and difference
  mechanism.
\newblock In \emph{CVPR Workshops}, 2021.

\bibitem[Dangovski et~al.(2022)Dangovski, Jing, Loh, Han, Srivastava, Cheung,
  Agrawal, and Soljai{\'c}]{Dangovski2021EquivariantCL}
Rumen Dangovski, Li~Jing, Charlotte Loh, Seung-Jun Han, Akash Srivastava, Brian
  Cheung, Pulkit Agrawal, and Marin Soljai{\'c}.
\newblock Equivariant contrastive learning.
\newblock In \emph{ICLR}, 2022.

\bibitem[Fischer et~al.(2018)Fischer, Chang, and Demiris]{Fischer2018RTGENERE}
Tobias Fischer, Hyung~Jin Chang, and Y.~Demiris.
\newblock Rt-gene: Real-time eye gaze estimation in natural environments.
\newblock In \emph{ECCV}, 2018.

\bibitem[Ghosh et~al.(2022)Ghosh, Hayat, Dhall, and Knibbe]{Ghosh2021MTGLSMG}
Shreya Ghosh, Munawar Hayat, Abhinav Dhall, and Jarrod Knibbe.
\newblock Mtgls: Multi-task gaze estimation with limited supervision.
\newblock In \emph{WACV}, 2022.

\bibitem[Gidaris et~al.(2018)Gidaris, Singh, and
  Komodakis]{gidaris2018unsupervised}
Spyros Gidaris, Praveer Singh, and Nikos Komodakis.
\newblock Unsupervised representation learning by predicting image rotations.
\newblock In \emph{ICLR}, 2018.

\bibitem[Grill et~al.(2020)Grill, Strub, Altch\'{e}, Tallec, Richemond,
  Buchatskaya, Doersch, Avila~Pires, Guo, Gheshlaghi~Azar, Piot, kavukcuoglu,
  Munos, and Valko]{Grill2020BootstrapYO}
Jean-Bastien Grill, Florian Strub, Florent Altch\'{e}, Corentin Tallec, Pierre
  Richemond, Elena Buchatskaya, Carl Doersch, Bernardo Avila~Pires, Zhaohan
  Guo, Mohammad Gheshlaghi~Azar, Bilal Piot, koray kavukcuoglu, Remi Munos, and
  Michal Valko.
\newblock Bootstrap your own latent - a new approach to self-supervised
  learning.
\newblock In \emph{NeurIPS}, 2020.

\bibitem[He et~al.(2016)He, Zhang, Ren, and Sun]{He2016DeepRL}
Kaiming He, X.~Zhang, Shaoqing Ren, and Jian Sun.
\newblock Deep residual learning for image recognition.
\newblock In \emph{CVPR}, 2016.

\bibitem[He et~al.(2020)He, Fan, Wu, Xie, and Girshick]{He2020MomentumCF}
Kaiming He, Haoqi Fan, Yuxin Wu, Saining Xie, and Ross~B. Girshick.
\newblock Momentum contrast for unsupervised visual representation learning.
\newblock In \emph{CVPR}, 2020.

\bibitem[Jaiswal et~al.(2020)Jaiswal, Babu, Zadeh, Banerjee, and
  Makedon]{Jaiswal2020ASO}
Ashish Jaiswal, Ashwin~Ramesh Babu, Mohammad~Zaki Zadeh, Debapriya Banerjee,
  and Fillia Makedon.
\newblock A survey on contrastive self-supervised learning.
\newblock In \emph{ArXiv}, 2020.

\bibitem[Kellnhofer et~al.(2019)Kellnhofer, Recasens, Stent, Matusik, and
  Torralba]{Kellnhofer2019Gaze360PU}
Petr Kellnhofer, Adri{\`a} Recasens, Simon Stent, W.~Matusik, and A.~Torralba.
\newblock Gaze360: Physically unconstrained gaze estimation in the wild.
\newblock In \emph{ICCV}, 2019.

\bibitem[Kingma and Ba(2015)]{adam-kingma}
Diederick~P Kingma and Jimmy Ba.
\newblock Adam: A method for stochastic optimization.
\newblock In \emph{ICLR}, 2015.

\bibitem[Kothari et~al.(2021)Kothari, Mello, Iqbal, Byeon, Park, and
  Kautz]{Kothari2021WeaklySupervisedPU}
Rakshit Kothari, Shalini~De Mello, Umar Iqbal, Wonmin Byeon, Seonwook Park, and
  Jan Kautz.
\newblock Weakly-supervised physically unconstrained gaze estimation.
\newblock In \emph{CVPR}, 2021.

\bibitem[Krafka et~al.(2016)Krafka, Khosla, Kellnhofer, Kannan, Bhandarkar,
  Matusik, and Torralba]{Krafka2016EyeTF}
K.~Krafka, A.~Khosla, Petr Kellnhofer, Harini Kannan, S.~Bhandarkar,
  W.~Matusik, and A.~Torralba.
\newblock Eye tracking for everyone.
\newblock In \emph{CVPR}, 2016.

\bibitem[Liu et~al.(2021)Liu, Zhang, Hou, Mian, Wang, Zhang, and Tang]{9462394}
Xiao Liu, Fanjin Zhang, Zhenyu Hou, Li~Mian, Zhaoyu Wang, Jing Zhang, and Jie
  Tang.
\newblock Self-supervised learning: Generative or contrastive.
\newblock In \emph{IEEE Transactions on Knowledge and Data Engineering}, 2021.

\bibitem[Loshchilov and Hutter(2017)]{Loshchilov2017SGDRSG}
Ilya Loshchilov and Frank Hutter.
\newblock {SGDR}: Stochastic gradient descent with warm restarts.
\newblock In \emph{ICLR}, 2017.

\bibitem[Misra and van~der Maaten(2020)]{Misra2020SelfSupervisedLO}
Ishan Misra and Laurens van~der Maaten.
\newblock Self-supervised learning of pretext-invariant representations.
\newblock In \emph{CVPR}, 2020.

\bibitem[Mora et~al.(2014)Mora, Monay, and Odobez]{Mora2014EYEDIAPAD}
Kenneth Alberto~Funes Mora, Florent Monay, and Jean-Marc Odobez.
\newblock Eyediap: a database for the development and evaluation of gaze
  estimation algorithms from rgb and rgb-d cameras.
\newblock In \emph{ETRA}, 2014.

\bibitem[Noroozi and Favaro(2016)]{Noroozi2016UnsupervisedLO}
Mehdi Noroozi and Paolo Favaro.
\newblock Unsupervised learning of visual representations by solving jigsaw
  puzzles.
\newblock In \emph{ECCV}, 2016.

\bibitem[Palmero et~al.(2018)Palmero, Selva, Bagheri, and
  Escalera]{Palmero2018RecurrentCF}
Cristina Palmero, Javier Selva, Mohammad~Ali Bagheri, and Sergio Escalera.
\newblock Recurrent cnn for 3d gaze estimation using appearance and shape cues.
\newblock In \emph{BMVC}, 2018.

\bibitem[Park et~al.(2018)Park, Spurr, and Hilliges]{Park2018DeepPG}
Seonwook Park, Adrian Spurr, and Otmar Hilliges.
\newblock Deep pictorial gaze estimation.
\newblock In \emph{ECCV}, 2018.

\bibitem[Parkhi et~al.(2015)Parkhi, Vedaldi, and Zisserman]{Parkhi15}
Omkar~M. Parkhi, Andrea Vedaldi, and Andrew Zisserman.
\newblock Deep face recognition.
\newblock In \emph{BMVC}, 2015.

\bibitem[Pathak et~al.(2016)Pathak, Kr{\"a}henb{\"u}hl, Donahue, Darrell, and
  Efros]{Pathak2016ContextEF}
Deepak Pathak, Philipp Kr{\"a}henb{\"u}hl, Jeff Donahue, Trevor Darrell, and
  Alexei~A. Efros.
\newblock Context encoders: Feature learning by inpainting.
\newblock In \emph{CVPR}, 2016.

\bibitem[Spurr et~al.(2021)Spurr, Dahiya, Wang, Zhang, and
  Hilliges]{spurr2021peclr}
Adrian Spurr, Aneesh Dahiya, Xi~Wang, Xucong Zhang, and Otmar Hilliges.
\newblock Self-supervised 3d hand pose estimation from monocular rgb via
  contrastive learning.
\newblock In \emph{ICCV}, 2021.

\bibitem[Sun et~al.(2021)Sun, Zeng, Shan, and Chen]{Sun_2021_ICCV}
Yunjia Sun, Jiabei Zeng, Shiguang Shan, and Xilin Chen.
\newblock Cross-encoder for unsupervised gaze representation learning.
\newblock In \emph{ICCV}, 2021.

\bibitem[Worrall et~al.(2017)Worrall, Garbin, Turmukhambetov, and
  Brostow]{Worrall2017InterpretableTW}
Daniel~E. Worrall, Stephan~J. Garbin, Daniyar Turmukhambetov, and Gabriel~J.
  Brostow.
\newblock Interpretable transformations with encoder-decoder networks.
\newblock In \emph{ICCV}, 2017.

\bibitem[Wu et~al.(2022)Wu, Li, Liu, Huang, and Wang]{9717236}
Yong Wu, Gongyang Li, Zhi Liu, Mengke Huang, and Yang Wang.
\newblock Gaze estimation via modulation-based adaptive network with auxiliary
  self-learning.
\newblock In \emph{IEEE Transactions on Circuits and Systems for Video
  Technology}, 2022.

\bibitem[Xie et~al.(2021)Xie, Ding, Wang, Zhan, Xu, Sun, Li, and
  Luo]{xie2021detco}
Enze Xie, Jian Ding, Wenhai Wang, Xiaohang Zhan, Hang Xu, Peize Sun, Zhenguo
  Li, and Ping Luo.
\newblock Detco: Unsupervised contrastive learning for object detection.
\newblock In \emph{ICCV}, 2021.

\bibitem[Xie et~al.(2022)Xie, Wen, Lau, Rehman, and Shen]{Xie_2022_CVPR}
Yuyang Xie, Jianhong Wen, Kin~Wai Lau, Yasar Abbas~Ur Rehman, and Jiajun Shen.
\newblock What should be equivariant in self-supervised learning.
\newblock In \emph{CVPR Workshops}, 2022.

\bibitem[You et~al.(2017)You, Gitman, and Ginsburg]{You2017LargeBT}
Yang You, Igor Gitman, and Boris Ginsburg.
\newblock Large batch training of convolutional networks.
\newblock In \emph{ArXiv}, 2017.

\bibitem[Yu and Odobez(2020)]{Yu2020UnsupervisedRL}
Yuechen Yu and Jean-Marc Odobez.
\newblock Unsupervised representation learning for gaze estimation.
\newblock In \emph{CVPR}, 2020.

\bibitem[Zhang et~al.(2016)Zhang, Isola, and Efros]{zhang2016colorful}
Richard Zhang, Phillip Isola, and Alexei~A Efros.
\newblock Colorful image colorization.
\newblock In \emph{ECCV}, 2016.

\bibitem[Zhang et~al.(2015)Zhang, Sugano, Fritz, and Bulling]{7299081}
Xucong Zhang, Yusuke Sugano, Mario Fritz, and Andreas Bulling.
\newblock Appearance-based gaze estimation in the wild.
\newblock In \emph{CVPR}, 2015.

\bibitem[Zhang et~al.(2017)Zhang, Sugano, Fritz, and Bulling]{zhang2017s}
Xucong Zhang, Yusuke Sugano, Mario Fritz, and Andreas Bulling.
\newblock It’s written all over your face: Full-face appearance-based gaze
  estimation.
\newblock In \emph{CVPR Workshops}, 2017.

\bibitem[Zhang et~al.(2018)Zhang, Sugano, and Bulling]{Zhang2018RevisitingDN}
Xucong Zhang, Yusuke Sugano, and Andreas Bulling.
\newblock Revisiting data normalization for appearance-based gaze estimation.
\newblock In \emph{ETRA}, 2018.

\bibitem[Zhang et~al.(2019)Zhang, Sugano, Fritz, and
  Bulling]{Zhang2019MPIIGazeRD}
Xucong Zhang, Yusuke Sugano, Mario Fritz, and Andreas Bulling.
\newblock Mpiigaze: Real-world dataset and deep appearance-based gaze
  estimation.
\newblock In \emph{TPAMI}, volume~41, pages 162--175, 2019.

\bibitem[Zhang et~al.(2020)Zhang, Park, Beeler, Bradley, Tang, and
  Hilliges]{Zhang2020ETHXGaze}
Xucong Zhang, Seonwook Park, Thabo Beeler, Derek Bradley, Siyu Tang, and Otmar
  Hilliges.
\newblock Eth-xgaze: A large scale dataset for gaze estimation under extreme
  head pose and gaze variation.
\newblock In \emph{ECCV}, 2020.

\end{thebibliography}
\end{document}